\documentclass[10pt,twocolumn,letterpaper]{article}

\usepackage[pagenumbers]{cvpr}              %

\newcommand{\red}[1]{{\color{red}#1}}

\usepackage{colortbl}
\usepackage{array}
\usepackage{tabularx}
\newcolumntype{Y}{>{\raggedright\arraybackslash}X}
\definecolor{oursbg}{HTML}{E8F3FF}

\definecolor{cvprblue}{rgb}{0.21,0.49,0.74}
\usepackage[pagebackref,breaklinks,colorlinks,allcolors=cvprblue]{hyperref}

\def\confName{CVPR}
\def\confYear{2026}

\title{Learning to Hear by Seeing: It's Time for Vision Language Models to Understand Artistic Emotion from Sight and Sound}

\author{
Dengming Zhang$^{1}$\quad
Weitao You$^{1}$$^*$\quad
Jingxiong Li$^{1}$\quad
Weishen Lin$^{1}$\quad
Wenda Shi$^{2}$\\
Xue Zhao$^{1}$\quad
Heda Zuo$^{1}$\quad
Junxian Wu$^{1}$\quad
Lingyun Sun$^{1}$\vspace{0.5em} \\ 
$^{1}$Zhejiang University \quad
$^{2}$The Hong Kong Polytechnic University \quad
\textsuperscript{$^*$}Corresponding author
}

\begin{document}
\maketitle
\begin{abstract}
Emotion understanding is critical for making Large Language Models (LLMs) more general, reliable, and aligned with humans. Art conveys emotion through the joint design of visual and auditory elements, yet most prior work is human-centered or single-modality, overlooking the emotion intentionally expressed by the artwork. Meanwhile, current Audio-Visual Language Models (AVLMs) typically require large-scale audio pretraining to endow Visual Language Models (VLMs) with hearing, which limits scalability. We present Vision Anchored Audio-Visual Emotion LLM (VAEmotionLLM), a two-stage framework that teaches a VLM to hear by seeing with limited audio pretraining and to understand emotion across modalities. In Stage 1, Vision-Guided Audio Alignment (VG-Align) distills the frozen visual pathway into a new audio pathway by aligning next-token distributions of the shared LLM on synchronized audio-video clips, enabling hearing without a large audio dataset. In Stage 2, a lightweight Cross-Modal Emotion Adapter (EmoAdapter), composed of the Emotion Enhancer and the Emotion Supervisor, injects emotion-sensitive residuals and applies emotion supervision to enhance cross-modal emotion understanding. We also construct ArtEmoBenchmark, an art-centric emotion benchmark that evaluates content and emotion understanding under audio-only, visual-only, and audio-visual inputs. VAEmotionLLM achieves state-of-the-art results on ArtEmoBenchmark, outperforming audio-only, visual-only, and audio-visual baselines. Ablations show that the proposed components are complementary.
\end{abstract}
    
\vspace{-0.5em}
\section{Introduction}

Emotion understanding is a central capability on the path toward more general and trustworthy Artificial Intelligence (AI) \cite{feng2024far, shou2025multimodal, zhou2024memo, zhang2024affective}.
As artworks are important carriers of human emotions, correctly understanding the emotions expressed in art is a necessary step for current AI toward Artificial General Intelligence (AGI).
Movies, music, and other forms of artwork often encode rich emotional signals through visual and auditory information, including abstract and metaphorical expression.
However, existing emotion-capable large language models (LLMs) primarily focus on human-centered emotion recognition (such as identifying how characters feel from faces, actions, or dialogue \cite{cheng2024emotion, zhao2025humanomni}) rather than recognizing the emotions conveyed by the artwork itself.
As a result, the ability to understand the emotional intent of artworks remains a gap in current LLMs.

To bridge this gap, prior work has introduced various Multimodal LLMs (MLLMs) that attempt to interpret artworks from visual \cite{yuan2023artgpt, hayashi2024artwork, li2025eyesee, ozaki2024towards} or auditory \cite{yuan2024chatmusician, liu2023m, liu2024music} cues. However, these approaches typically focus on a single modality and lack robust emotional perception of the artwork, limiting their practical value.
In fact, the emotions expressed in artworks are conveyed simultaneously through both visual and auditory modalities. Relying on only one leads to partial and limited emotional understanding.
For example, directors often use background music to create emotional contrast or enhance the mood of a scene. The same visuals paired with different soundtracks can evoke entirely different emotions.
Therefore, current MLLMs must address two challenges: \emph{Audio-Visual Modality Completeness}, meaning simultaneous support for visual and auditory inputs, and \emph{Cross-Modal Emotion Understanding}, meaning strong emotional perception and cross-modal understanding that yield emotions reflecting their joint influence rather than a simple average or single-modality bias.

Regarding Audio-Visual Modality Completeness, although recent Audio-Visual Language Models (AVLMs), such as Qwen2.5-Omni \cite{xu2025qwen2} and VideoLLaMA \cite{zhang2023video, cheng2024videollama}, report promising results, they typically require large-scale audio-visual pretraining to enable dual-modality input. This makes it difficult to rapidly upgrade many strong Visual Language Models (VLMs) \cite{bai2025qwen2_5vl, chen2024internvl, chen2024expanding} into AVLMs. Regarding Cross-Modal Emotion Understanding, most current AVLMs adopt separate encoders for vision and audio and are guided predominantly through instruction following. This design hampers Cross-Modal Emotion Understanding: models perform well on single-modality emotion cues yet struggle to infer holistic emotion in complex scenes where sound and imagery interact. 
AffectGPT \cite{lian2025affectgpt} addresses this by pre-fusing multimodal tokens to explicitly input into the LLM, enhancing multimodal integration. However, this approach still mainly relies on the LLM to model fused modality information, making it challenging for the model to fully grasp complex cross-modal emotions.

To address these two challenges, we propose Vision Anchored Audio-Visual Emotion LLM (VAEmotionLLM), a two-stage framework that teaches a strong VLM to hear by seeing and to understand emotion across modalities. In Stage 1, Vision-Guided Audio Alignment (VG-Align) aligns an audio encoder equipped with an audio adapter to the VLM's vision encoder on unlabeled synchronized audio-video clips, injecting auditory perception into the VLM. In Stage 2, a two-part Cross-Modal Emotion Adapter (EmoAdapter) introduces emotion-sensitive multimodal tokens and applies emotion supervision to strengthen cross-modal emotion understanding. To evaluate artistic emotion understanding, we propose the Art-centric Emotion Benchmark (ArtEmoBenchmark), which focuses on understanding the content and emotion of movies with background music, where emotion is crafted through the interplay of sight and sound. With minimal audio supervision, VAEmotionLLM achieves state-of-the-art results over audio-only, visual-only, and audio-visual baselines, and ablations confirm the complementary roles of our proposed components.

The main contributions are as follows:
\begin{itemize}
    \item VAEmotionLLM and ArtEmoBenchmark: we introduce a two-stage audio-visual emotion LLM and release an art-centric emotion benchmark that focuses on understanding the content and emotion of movies with background music across three input modalities.
    \item VG-Align: a vision-guided alignment that maps an audio pathway to the VLM visual token space using unlabeled synchronized audio-video clips, enabling audio comprehension without large-scale audio pretraining.
    \item EmoAdapter: a lightweight two-part adapter, Emotion Enhancer and Emotion Supervisor, that injects emotion into modality tokens and applies emotion supervision over audio, visual, and audio-visual joint paths to improve cross-modal emotion understanding.
\end{itemize}

\section{Related Work}
\subsection{MLLMs: From VLMs to AVLMs}
MLLMs have rapidly progressed along three lines that are most relevant to our setting: Visual Language Models (VLMs), Audio Language Models (ALMs), and Audio-Visual Language Models (AVLMs)~\cite{wang2024comprehensive, song2025bridge}. VLMs couple a strong visual encoder with a powerful LLM backbone through a learned projector and extensive multimodal instruction tuning, yielding robust visual understanding \cite{chen2024internvl, bai2025qwen2_5vl, chen2024expanding}. In parallel, recent ALMs focus on audio understanding, captioning, and even controllable generation, advancing audio question answering and description \cite{chu2023qwen, chu2024qwen2, yuan2024chatmusician}. Building on these foundations, AVLMs extend the interface to video and audio. VideoLLaMA~\cite{zhang2023video} and its successor~\cite{cheng2024videollama} integrate separate vision and audio encoders with an LLM and are trained with large-scale instruction data for video understanding and audio grounding. Qwen2.5-Omni \cite{xu2025qwen2} pursues a unified formulation that interleaves tokens from different modalities with time-aligned multimodal rotary position embeddings, improving cross-modal temporal modeling. Despite strong results, these AVLMs typically rely on substantial audio-visual pretraining and wide-ranging instruction data to endow models with dual-modality input, which makes rapid upgrading of strong VLMs into AVLMs costly. To address this, we introduce Vision-Guided Audio Alignment (VG-Align) that aligns new audio path with the vision encoder of a frozen VLM so the model can hear by seeing without large-scale audio pretraining.

\subsection{Emotion Understanding Based on LLMs}
Affective computing studies how to enable machines to perceive, understand, and communicate human emotions~\cite{zhang2024affective, shou2025multimodal} across text~\cite{gupta2024comprehensive, yang2025large, kumar2023analyzing}, audio~\cite{wang2025enhanced, zhang2025personalized, chang2024iiof}, and video~\cite{wei2024learning, dutta2025llm, sarvakar2023facial}. With the rise of AI~\cite{shi2025fonts,shi2025wordcon,shi2025generative,zhang2025expert}, many methods have been introduced to endow general models with emotional capabilities~\cite{yang2025mse,lian2025affectgpt, yang2024emollm, cheng2024emotion, huang2025emotion}. MSE-Adapter~\cite{yang2025mse} is a lightweight plugin that equips an LLM to perform Multimodal Sentiment Analysis and Emotion Recognition by fusing audio and visual cues, achieving strong label prediction accuracy while offering limited interpretability. Emotion-LLaMA~\cite{cheng2024emotion} goes further by integrating audio and recognizing subtle facial micro-expressions, enabling instruction-tuned multimodal emotion understanding, yet it remains primarily human-centric. These efforts effectively enhance emotion understanding but remain largely focused on human-centered emotion recognition. As a result, they struggle to accurately capture artistic emotion, which in artworks arises from the joint interplay of vision and sound at the level of the whole piece. We address this with VG-Align for vision-guided hearing and a lightweight EmoAdapter for cross-modal emotion understanding, integrated into a two-stage framework that learns to hear by seeing and to understand art across modalities.

\section{Method}

\subsection{Notation and terminology.}
We denote the frozen vision tower as $f_v$, the pretrained audio encoder as $e_a$, the audio adapter as $g_{\phi}$ with parameters $\phi$, and the shared LLM as $F_{\theta}$ with frozen weights $\theta$ unless noted. Inputs are $(x^v, x^a)$ for video and audio. The LLM consumes modality tokens $\mathbf{z}_m\in\mathbb{R}^{L_m\times d}$, where $m\in\{v,a,av\}$ indexes visual-only, audio-only, or audio-visual joint.

\subsection{Stage 1: Let VLM Hear with Eyes}
\label{subsec:stage1}

\begin{figure*}
  \centering
  \includegraphics[width=\linewidth]{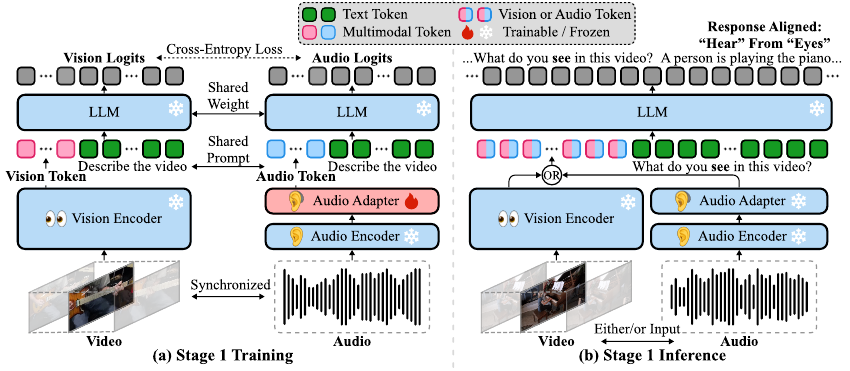}
  \caption{Vision-Guided Audio Alignment. Left: training distills a frozen visual pathway (teacher) into an audio pathway (student) by matching next-token distributions under a shared prompt while the VLM and the vision encoder are frozen. Right: after training, either video or audio can be fed to the shared LLM to produce aligned responses.}
  \label{fig:vg_align}
\end{figure*}

\textbf{Goal and intuition.}
Strong VLMs already know how to describe what they see. VG-Align teaches the VLM to hear by learning an audio pathway whose effect on the LLM matches that of the visual pathway, using unlabeled synchronized audio-video clips. Instead of forcing audio features to numerically equal vision features, we align the \emph{response distribution} of the shared LLM, so the model produces the same answers given audio as it would given video.

\textbf{Architecture.}
We keep the pretrained VLM frozen, including the vision encoder and the language model. For audio we adopt Audio Spectrogram Transformer (AST)~\cite{gong21b_interspeech} as the pretrained audio encoder and add a lightweight \emph{Audio Adapter} that reshapes its output to the visual token space used by the LLM. The adapter is explicit and simple. It uses two fully connected layers to reduce the AST sequence into the visual token length $L_v$, then two fully connected layers to project the channel dimension to the visual embedding dimension $d_v$. Given AST features of shape $[T_a, d_a]$ the adapter outputs $\mathbf{z}_a \in \mathbb{R}^{L_v\times d_v}$. We insert these audio tokens at positions reserved for video tokens and assign them the same positional encodings as the corresponding visual tokens by reusing the vision RoPE grid. All parameters of the LLM, vision encoder and audio encoder are frozen and only the audio adapter is trainable.

\textbf{Teacher-student distribution alignment.}
For each synchronized audio-video pair $(x^v, x^a)$ and a prompt $s$ randomly sampled from a predefined pool, we treat the \emph{Vision Encoder + LLM} as the teacher and the \emph{Audio Encoder + Audio Adapter + LLM} as the student. The teacher path encodes video to $\mathbf{z}_v=f_v(x^v)$ and produces teacher logits $\ell_v^{(t)}=F_\theta(s,\mathbf{z}_v)^{(t)}$ with the frozen LLM. The student path maps audio to $\mathbf{z}_a=g_\phi(e_a(x^a))$ and produces student logits $\ell_a^{(t)}=F_\theta(s,\mathbf{z}_a)^{(t)}$ with the \emph{same} frozen LLM. We optimize the student to match the teacher by a soft cross-entropy loss
\begin{equation}
\label{eq:vg_loss}
\mathcal{L}_{\text{VG}}(\phi)=\frac{1}{T}\sum_{t=1}^{T}\,\mathrm{CE}\big(\sigma(\ell_v^{(t)}/\tau),\, \sigma(\ell_a^{(t)}/\tau)\big),
\end{equation}
where $\sigma$ is softmax, $\tau>0$ is a temperature, $T$ is the number of assistant tokens, and $\mathrm{CE}(p,q)\triangleq -\sum_i p_i\log q_i$ is the cross-entropy between two distributions. This objective uses no human labels and transfers knowledge anchored in the vision encoder to the audio pathway, enabling the model to learn to hear from its eyes.

\textbf{Why logit alignment instead of feature alignment.}
An intuitive alternative is to force $\mathbf{z}_a$ to equal $\mathbf{z}_v$ via $\ell_2$ loss. In practice this is suboptimal. The token to logit mapping of $F_\theta$ is many to one, which means exact equality of features is unnecessary and can even harm learning, because many distinct $\tilde{\mathbf{z}}$ induce the same next token distribution as $\mathbf{z}_v$. The audio and visual encoders also have different geometries, so strict feature coincidence is brittle even with an adapter. What ultimately matters is the conditional distribution that drives reasoning, $p_\theta(y_t\,|\,s,\mathbf{z},y_{<t})$. Minimizing soft cross-entropy between teacher and student logits aligns this distribution since $\mathrm{CE}(p_v,p_a)=\mathrm{KL}(p_v\Vert p_a)+H(p_v)$ and the entropy term is constant, hence the minimum is achieved when $p_a=p_v$ regardless of how $\mathbf{z}_a$ differs from $\mathbf{z}_v$. Empirically we observed better stability and stronger performance with this response-level alignment, as shown in Table~\ref{tab:air_benchmark}.

\subsection{Stage 2: Cross-Modal Emotion Adapter}
\label{subsec:stage2}

\begin{figure}
  \centering
  \includegraphics[width=\linewidth]{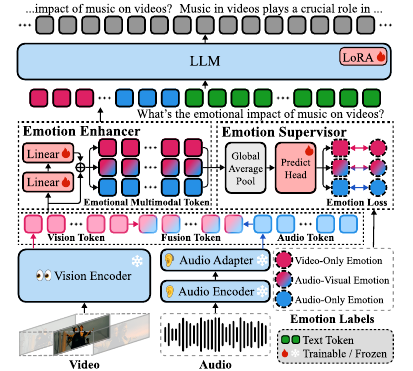}
  \caption{Architecture of the EmoAdapter (Stage 2). The \emph{Emotion Enhancer} injects lightweight residuals into audio and video tokens to form emotion-sensitive multimodal tokens. The \emph{Emotion Supervisor} aggregates tokens and predicts V-A labels for three pathways (audio-only, visual-only, and audio-visual joint).}
  \label{fig:stage_2}
\end{figure}

Stage 1 enables the model to hear from its eyes, but it accepts only unimodal input (audio or video) and still lacks cross-modal emotion understanding. To address this limitation, Stage 2 introduces the \emph{Cross-Modal Emotion Adapter} (EmoAdapter), a minimal add-on composed of two parts: \emph{Emotion Enhancer} and \emph{Emotion Supervisor}. 
The Emotion Enhancer injects modality-agnostic emotion information through modality-shared layers via residual connections to produce emotion-sensitive tokens, while the Emotion Supervisor aggregates tokens and predicts emotion labels for three pathways (audio-only, visual-only, and audio-visual joint) under emotion label supervision. Here, we use the continuous Valence-Arousal (V-A) representation~\cite{russell1980circumplex} as the emotion label, where a V-A label denotes two real-valued scalars (valence, arousal) that describe emotion polarity and intensity, respectively.

\textbf{Emotion Enhancer.}
To make unimodal tokens emotion-sensitive while keeping the encoders frozen, we introduce a modality-shared lightweight enhancer that injects modality-agnostic emotion information into multimodal tokens via residual connections.
Let $\mathbf{z}_m\in\mathbb{R}^{L_m\times d}$ be the aligned Stage~1 tokens for modality $m\in\{v,a,av\}$. The enhancer is a two-layer residual MLP shared across all modalities and applied independently to each token position:
\begin{equation}
\label{eq:enhancer}
\mathrm{Enh}(\mathbf{z}_m)
= \mathbf{z}_m+\underbrace{W_2\,\sigma\!\big(W_1\,\mathrm{LN}(\mathbf{z}_m)\big)}_{\triangleq\,E_{\phi}(\mathbf{z}_m)},
\end{equation}
where $\mathrm{LN}$ denotes layer normalization~\cite{ba2016layer}, $\sigma$ is the GELU nonlinearity, and $W_1,W_2$ are learnable linear projections. We denote the parameters of the enhancer by $\phi=\{W_1,W_2\}$, and write $E_{\phi}:\mathbb{R}^{L_m\times d}\rightarrow\mathbb{R}^{L_m\times d}$ for the corresponding residual mapping. Equation~\eqref{eq:enhancer} thus injects an emotion perturbation $E_{\phi}(\mathbf{z}_m)$ in the shared token space for all modalities. Sharing forces the same set of residual directions to be useful for visual-only, audio-only, and audio-visual joint pathways, implicitly tying how the model attends to rhythm, timbre, and scene composition.
The enhanced tokens $\widetilde{\mathbf{z}}_a=\mathrm{Enh}(\mathbf{z}_a)$ and $\widetilde{\mathbf{z}}_v=\mathrm{Enh}(\mathbf{z}_v)$ are fed to the LLM for understanding the emotion from the respective modality or their combination. 

\textbf{Emotion Supervisor.}
While the Emotion Enhancer produces emotion-sensitive tokens that implicitly fuse multimodal cues, it is still difficult for the LLM to discover the correct emotion information without explicit supervision. Therefore, we propose the Emotion Supervisor to add explicit emotion supervision that guides the model to capture the correct emotion information.
For emotion supervision, we aggregate tokens by global average pooling and predict continuous V-A labels. Define summaries
\begin{equation}
\mathbf{s}_m=\frac{1}{L_m}\sum_{t=1}^{L_m}\widetilde{\mathbf{z}}_m^{(t)},\qquad m\in\{v,a,av\},
\end{equation}
and a fused sequence by concatenation $\mathbf{z}_{av}=[\mathbf{z}_a;\mathbf{z}_v]$ followed by the same enhancer to obtain the emotional fused tokens $\widetilde{\mathbf{z}}_{av}$ and fused summaries $\mathbf{s}_{av}$. The Emotion Supervisor, parameterized by $\psi$ and shared across modalities, produces predicted V-A labels
\begin{equation}
\hat{\mathbf{y}}_m=g_\psi(\mathbf{s}_m)\in\mathbb{R}^2,\qquad m\in\{v,a,av\},
\end{equation}
which are two floats (valence, arousal). Because the enhancer is shared and the fused path reuses it, gradients from $\hat{\mathbf{y}}_{av}$ couple the audio and video streams and align them in an emotion-aware subspace.

\textbf{Learning objective.}
Let $\mathcal{M}\subseteq\{a,v,av\}$ be the set of available labels for a sample. We minimize
\begin{equation}
\label{eq:emo_loss}
\mathcal{L}_{\text{emo}}=\frac{1}{|\mathcal{M}|}\sum_{m\in\mathcal{M}}\lVert\hat{\mathbf{y}}_m-\mathbf{y}_m\rVert_2^2,
\end{equation}
and jointly optimize the conversational language loss $\mathcal{L}_{\text{LM}}$ from the LLM on the same training instance,
\begin{equation}
\label{eq:total_loss}
\mathcal{L}=\mathcal{L}_{\text{LM}}+\lambda\,\mathcal{L}_{\text{emo}}.
\end{equation}
In practice we fine-tune only the Emotion Enhancer, the Emotion Supervisor, and the Low-Rank Adaptation (LoRA)~\cite{hu2022lora} inside the LLM, while keeping the vision encoder, audio encoder, and the audio adapter frozen. This design greatly reduces computational cost.

\textbf{Why this improves cross-modal emotion.}
Equation~\eqref{eq:enhancer} learns a shared residual basis $E_{\phi}(\mathbf{z}_m)$ that perturbs tokens in a common emotion subspace for all modalities. Because the fused $av$ pathway is included in \eqref{eq:emo_loss}, this basis is required to explain emotion signals that only emerge when audio and vision interact. Sharing the enhancer and supervisor means that gradients from every supervised pathway accumulate on the same parameters:
\begin{equation}
\nabla_{\phi}\,\mathcal{L}_{\text{emo}}=\sum_{m\in\mathcal{M}}\frac{\partial\mathcal{L}_{\text{emo}}}{\partial\hat{\mathbf{y}}_m}\frac{\partial g_\psi(\mathbf{s}_m)}{\partial \mathbf{s}_m}\frac{\partial\mathbf{s}_m}{\partial\widetilde{\mathbf{z}}_m}\frac{\partial E_{\phi}(\mathbf{z}_m)}{\partial\phi},
\end{equation}
and the supervisor itself is updated by
\begin{equation}
\nabla_{\psi}\,\mathcal{L}_{\text{emo}}=\sum_{m\in\mathcal{M}}\frac{\partial\mathcal{L}_{\text{emo}}}{\partial\hat{\mathbf{y}}_m}\frac{\partial g_\psi(\mathbf{s}_m)}{\partial \psi}.
\end{equation}
In words, the fused branch teaches the shared enhancer and supervisor to encode cross-modal cues (e.g., rhythm aligned with motion or music contradicting the scene), and the unimodal branches reuse the same emotion directions on audio-only or video-only tokens. Consequently, cross-modal emotion information is already aligned in the shared token space before reaching the LLM, which reduces the pressure on the LLM to learn cross-modal emotion alignment from scratch and enables more accurate cross-modal emotion judgments, as confirmed in Table~\ref{tab:emotion_benchmark}.

\subsection{Implementation Details}
\label{subsec:impl}
We instantiate VAEmotionLLM on a Qwen2.5-VL-7B~\cite{bai2025qwen2_5vl} backbone and use AST~\cite{gong21b_interspeech} as the audio encoder.
In Stage~1, we freeze all parameters except the audio adapter and set the temperature to $\tau = 1$.
In Stage~2, we apply LoRA to the LLM attention projections (q, k, v, o layers) while keeping the vision encoder, audio encoder, and audio adapter frozen; only the Emotion Enhancer, Emotion Supervisor, and LoRA parameters are updated.
The training objective is $\mathcal{L}=\mathcal{L}_{\text{LM}}+\lambda\,\mathcal{L}_{\text{emo}}$ with $\lambda=1$, where the emotion loss is computed over available modalities and missing labels are masked out.
We optimize the model with AdamW~\cite{loshchilov2017decoupled} using a learning rate of $1\times10^{-5}$. Input videos are resized to $512\times512$ and audio is resampled to 16kHz. All experiments are conducted on 8 $\times$ RTX 4090 GPUs.

\begin{table*}[!h]
    \centering
    \caption{AIR-Bench: performance across speech, acoustic, and music tasks. Columns are grouped by audio dataset hours: large-scale ($>$100k~h) vs. small-scale ($<$100k~h). Best within each group is in bold and second best is underlined. All numbers are percentages.}
    \label{tab:air_benchmark}
    \begingroup
    \setlength{\tabcolsep}{3pt}
    \renewcommand{\arraystretch}{1.08}
    \footnotesize
    \begin{tabular*}{\linewidth}{@{\extracolsep{\fill}}p{4.2cm} *{2}{c} | *{5}{c}}
        \toprule
        & \multicolumn{2}{c}{\textbf{Large-scale} ($>$100k h, \%)} & \multicolumn{5}{c}{\textbf{Small-scale} ($<$100k h, \%)} \\
        \cmidrule(lr){2-3} \cmidrule(lr){4-8}
        \textbf{Categories} & Qwen2-Audio  & Qwen-Audio  & Ours  & Ours$^{\dagger}$  & Next-GPT  & BLSP  & SpeechGPT  \\
        \midrule
        Speech Grounding & \underline{25.60} & \textbf{56.10} & \underline{28.13} & 27.11 & 25.40 & 25.00 & \textbf{28.80} \\
        Spoken Language Identification & \underline{82.76} & \textbf{92.80} & \textbf{40.80} & 35.90 & 23.70 & 30.80 & \underline{39.60} \\
        Speaker Gender Recognition & \textbf{70.01} & \underline{67.20} & \underline{37.58} & 28.79 & \textbf{57.00} & 33.20 & 29.20 \\
        Emotion Recognition & \textbf{53.49} & \underline{43.20} & \underline{34.97} & 32.00 & 25.70 & 27.40 & \textbf{37.60} \\
        Speaker Age Prediction & \underline{29.48} & \textbf{36.00} & 43.30 & 38.40 & \textbf{62.40} & \underline{51.20} & 20.40 \\
        Speech Entity Recognition & \textbf{87.90} & \underline{71.20} & 32.90 & \underline{36.11} & 26.10 & \textbf{37.20} & 35.90 \\
        Intent Classification & \textbf{85.55} & \underline{77.80} & 27.10 & 23.69 & 25.60 & \textbf{46.60} & \underline{45.80} \\
        Speaker Number Verification & \textbf{51.77} & \underline{35.30} & \textbf{39.30} & 32.30 & 25.40 & 28.10 & \underline{32.60} \\
        Synthesized Voice Detection & \underline{37.89} & \textbf{48.30} & \underline{47.20} & 44.10 & 30.80 & \textbf{50.00} & 39.20 \\
        Audio Grounding & \textbf{48.62} & \underline{23.90} & 29.35 & 26.16 & \textbf{62.20} & \underline{34.60} & 26.10 \\
        Vocal Sound Classification & \textbf{87.07} & \underline{84.90} & \underline{27.20} & 26.30 & 23.50 & \textbf{29.80} & 26.20 \\
        Acoustic Scene Classification & \underline{66.00} & \textbf{67.50} & \textbf{28.15} & \underline{28.10} & 24.10 & 25.20 & 23.70 \\
        Sound Question Answering & \textbf{71.29} & \underline{64.60} & \textbf{49.90} & \underline{44.25} & 18.80 & 36.10 & 33.90 \\
        Music Instruments Classification & \textbf{65.36} & \underline{59.10} & \textbf{32.00} & 25.85 & 24.30 & 22.80 & \underline{29.10} \\
        Music Genre Classification & \textbf{76.80} & \underline{71.20} & \textbf{30.10} & 26.65 & 28.10 & 26.10 & \underline{29.30} \\
        Music Note Analysis (Pitch) & \underline{27.60} & \textbf{28.60} & \textbf{28.23} & \underline{27.66} & 25.10 & 23.50 & 24.10 \\
        Music Note Analysis (Velocity) & \textbf{25.98} & \underline{25.40} & \textbf{27.20} & 24.90 & 23.10 & 24.90 & \underline{25.20} \\
        Music Question Answering & \textbf{71.37} & \underline{48.20} & \textbf{57.62} & \underline{50.12} & 47.10 & 31.00 & 31.30 \\
        Music Emotion Detection & \textbf{49.35} & \underline{36.10} & \underline{28.90} & 27.80 & 25.40 & 28.30 & \textbf{29.70} \\
        \midrule
        Average & \textbf{60.02} & \underline{54.50} & \textbf{35.17}{\textcolor{red}{\textsuperscript{\scriptsize (+3.67)}}} & \underline{31.59}{\textcolor{red}{\textsuperscript{\scriptsize (+0.09)}}} & 31.50 & 31.40 & 30.00 \\
        \bottomrule
    \end{tabular*}
    \par\vspace{.25em}\footnotesize $^{\dagger}$ Ours uses feature align, replacing logits align.
    \endgroup
    \vspace{-0.5em}
\end{table*}

\begin{table*}[!h]
    \centering
    \caption{Performance on ArtEmoBenchmark.}
    \label{tab:emotion_benchmark}
    \begingroup
    \setlength{\tabcolsep}{1.5pt}
    \renewcommand{\arraystretch}{1.05}
    \footnotesize
    \begin{tabular*}{\linewidth}{@{\extracolsep{\fill}}p{2.6cm}*{16}{c}}
        \toprule
        & \multicolumn{4}{c}{\textbf{Audio (A-only, \%)}} & \multicolumn{4}{c}{\textbf{Video (V-only, \%)}} & \multicolumn{4}{c}{\textbf{Audio-Visual (AV, joint, \%)}} & \multicolumn{4}{c}{\textbf{Averages (\%)}} \\
        \cmidrule(lr){2-5} \cmidrule(lr){6-9} \cmidrule(lr){10-13} \cmidrule(lr){14-17}
        \textbf{Model} & \textbf{{OC}} & \textbf{{OE}} & \textbf{{SC}} & \textbf{{SE}} & \textbf{{OC}} & \textbf{{OE}} & \textbf{{SC}} & \textbf{{SE}} & \textbf{{Sp-A}} & \textbf{{Sp-AV}} & \textbf{{Sp-V}} & \textbf{{OE}} & \textbf{{A}} & \textbf{{V}} & \textbf{{AV}} & \textbf{{All}} \\
        \midrule
        \rowcolor{gray!12}\multicolumn{17}{c}{\emph{Audio Language Models (ALMs)}} \\
        Qwen-Audio & 74.0 & 35.0 & 49.0 & 52.0 & -- & -- & -- & -- & -- & -- & -- & -- & 52.5 & -- & -- & -- \\
        Qwen2-Audio (7B) & 87.0 & 40.0 & 60.0 & 65.0 & -- & -- & -- & -- & -- & -- & -- & -- & 63.0 & -- & -- & -- \\
        \midrule
        \rowcolor{gray!12}\multicolumn{17}{c}{\emph{Visual Language Models (VLMs)}} \\
        Qwen2.5-VL (7B) & -- & -- & -- & -- & 84.0 & 55.0 & 69.0 & 81.0 & -- & -- & -- & -- & -- & 72.3 & -- & -- \\
        InternVL2.5 (8B) & -- & -- & -- & -- & 94.0 & 56.0 & \underline{81.0} & 79.0 & -- & -- & -- & -- & -- & 77.5 & -- & -- \\
        \midrule
        \rowcolor{gray!12}\multicolumn{17}{c}{\emph{Audio-Visual Language Models (AVLMs)}} \\
        Qwen2.5-Omni (7B) & \underline{88.0} & 50.0 & \textbf{72.0} & \underline{66.0} & 83.0 & 59.0 & 76.0 & 83.0 & 52.0 & 40.0 & 22.0 & \underline{53.0} & \underline{69.0} & 75.3 & 41.8 & 62.0 \\
        InteractiveOmni (8B) & 80.0 & 49.0 & 65.0 & 64.0 & 97.0 & 63.0 & \textbf{87.0} & \underline{85.0} & \underline{53.0} & \underline{42.0} & 25.0 & 46.0 & 64.5 & \underline{83.0} & 41.5 & \underline{63.0} \\
        VideoLLaMA2 (7B) & 77.0 & \underline{55.0} & 48.0 & 63.0 & \underline{98.0} & \underline{65.0} & \underline{81.0} & \underline{81.0} & 39.0 & 37.0 & 27.0 & 43.0 & 60.8 & 81.3 & 36.5 & 59.5 \\
        Emotion-LLaMA  & 36.0 & 33.0 & 27.0 & 39.0 & 56.0 & 30.0 & 46.0 & 59.0 & 40.0 & 35.0 & 32.0 & 42.0 & 33.8 & 47.8 & 37.2 & 39.6 \\
        AffectGPT & 69.0 & 36.0 & 48.0 & 53.0 & 57.0 & 42.0 & 63.0 & 81.0 & 48.0 & 36.0 & \underline{40.0} & 44.0 & 51.5 & 60.8 & \underline{42.0} & 51.4 \\
        \midrule
        \rowcolor{oursbg}Ours & \textbf{89.0} & \textbf{77.0} & \underline{70.0} & \textbf{72.0} & \textbf{100.0} & \textbf{74.0} & \textbf{87.0} & \textbf{90.0} & \textbf{60.0} & \textbf{53.0} & \textbf{42.0} & \textbf{59.0} & \textbf{77.0}{\textcolor{red}{\textsuperscript{\scriptsize (+8.0)}}} & \textbf{87.8}{\textcolor{red}{\textsuperscript{\scriptsize (+4.8)}}} & \textbf{53.5}{\textcolor{red}{\textsuperscript{\scriptsize (+11.5)}}} & \textbf{72.8}{\textcolor{red}{\textsuperscript{\scriptsize (+9.8)}}} \\
        \bottomrule
    \end{tabular*}
    \par\vspace{.25em}\scriptsize
    Abbreviations: OC=overall content questions; OE=overall emotion questions; SC=specific content questions; SE=specific emotion questions; Sp-A=audio-centric specific questions; 
    Sp-V=video-centric specific questions; Sp-AV=cross-modal specific questions(joint input);  A/V/AV/All=group and overall averages. All numbers are percentages. 
    \endgroup
    \vspace{-0.5em}
\end{table*}

\begin{table}[!h]
    \centering
    \caption{Ablation of VAEmotionLLM components on ArtEmoBenchmark.}
    \label{tab:ablation}
    \begingroup
    \setlength{\tabcolsep}{2pt}
    \renewcommand{\arraystretch}{1.05}
    \newcommand{\rot}[1]{\shortstack{\footnotesize #1}}
    \newcommand{\cmark}{\checkmark}
    \newcommand{\xmark}{$\times$}
    \footnotesize
    \begin{tabular*}{\linewidth}{@{\extracolsep{\fill}}cccccccc@{}}
        \toprule
        \multicolumn{4}{c}{\textbf{Components}} & \multicolumn{4}{c}{\textbf{Averages (\%)}} \\
        \cmidrule(lr){1-4} \cmidrule(lr){5-8}
        \rot{Audio\\Adapter} & \rot{LoRA} & \rot{Emotion\\Enhancer} & \rot{Emotion\\Supervisor} & \rot{A} & {\rot{V}} & {\rot{AV}} & {\rot{All}} \\
        \midrule
        \xmark & \xmark & \xmark & \xmark & -- & 72.3 & -- & -- \\
        \cmark & \xmark & \xmark & \xmark & 29.5 & 72.3 & -- & -- \\
        \cmark & \cmark & \xmark & \xmark & 45.8 & 72.0 & 37.5 & 51.8 \\
        \cmark & \cmark & \cmark & \xmark & \underline{52.3} & \underline{76.8} & \underline{43.8} & \underline{57.6} \\
        \cmark & \cmark & \xmark & \cmark & 46.8 & 73.0 & 38.3 & 52.7 \\
        \cmark & \cmark & \cmark & \cmark & \textbf{77.0} & \textbf{87.8} & \textbf{53.5} & \textbf{72.8} \\
        \bottomrule
    \end{tabular*}
    \par\vspace{.25em}\scriptsize
    A/V/AV/All denote group and overall averages. A dash indicates that the model lacks the required modality. All numbers are percentages.
    \endgroup
    \vspace{-0.5em}
\end{table}

\begin{table*}[!h]
    \centering
    \caption{Qualitative responses under different input modalities. The top block shows the input video frames and the audio description.}
    \label{tab:qualitative_av_responses}
    \begingroup
    \footnotesize
    {\centering
        \IfFileExists{figures/Example-1.png}{\includegraphics[width=.245\linewidth]{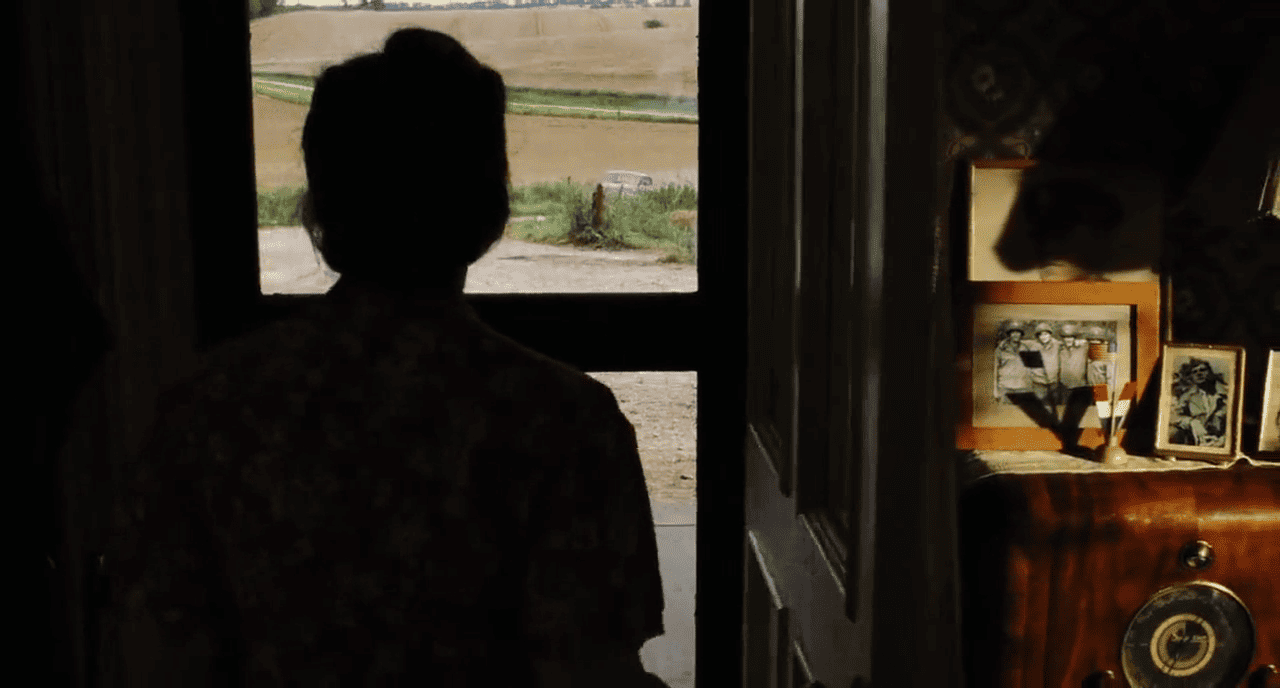}}{\fbox{\parbox[c][2.1cm][c]{.245\linewidth}{\centering Frame 1}}}%
        \hspace{.0065\linewidth}%
        \IfFileExists{figures/Example-2.png}{\includegraphics[width=.245\linewidth]{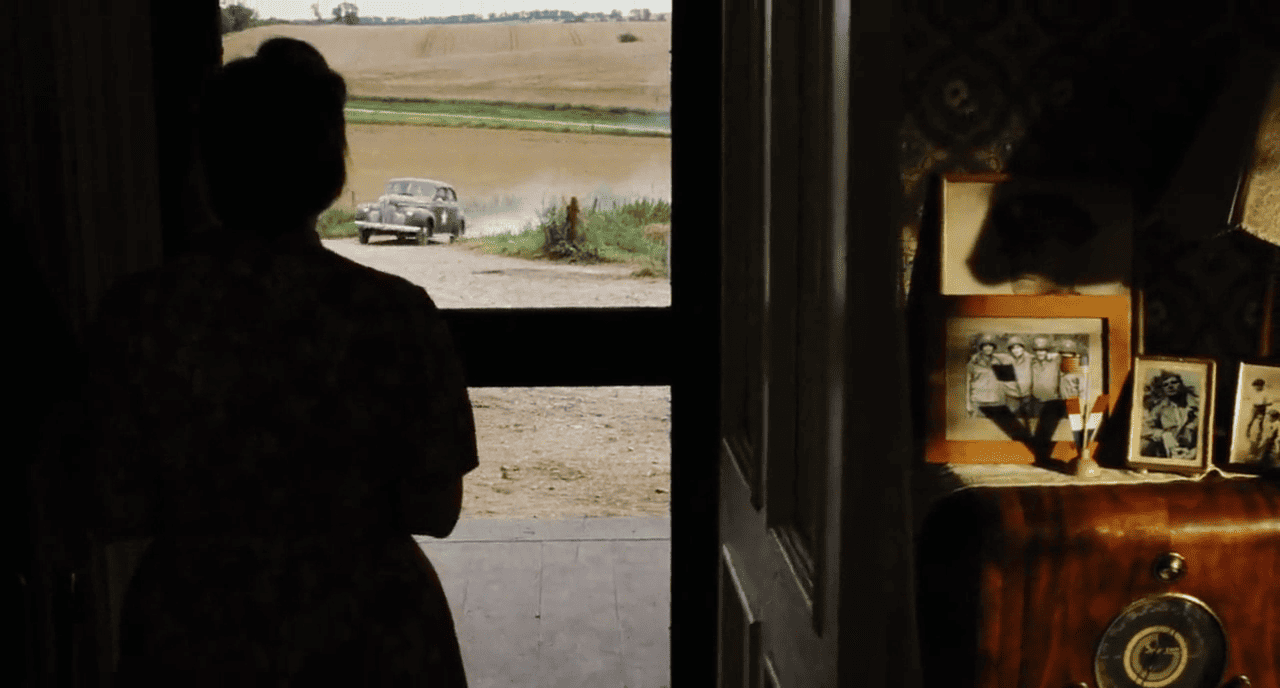}}{\fbox{\parbox[c][2.1cm][c]{.245\linewidth}{\centering Frame 2}}}%
        \hspace{.0065\linewidth}%
        \IfFileExists{figures/Example-3.png}{\includegraphics[width=.245\linewidth]{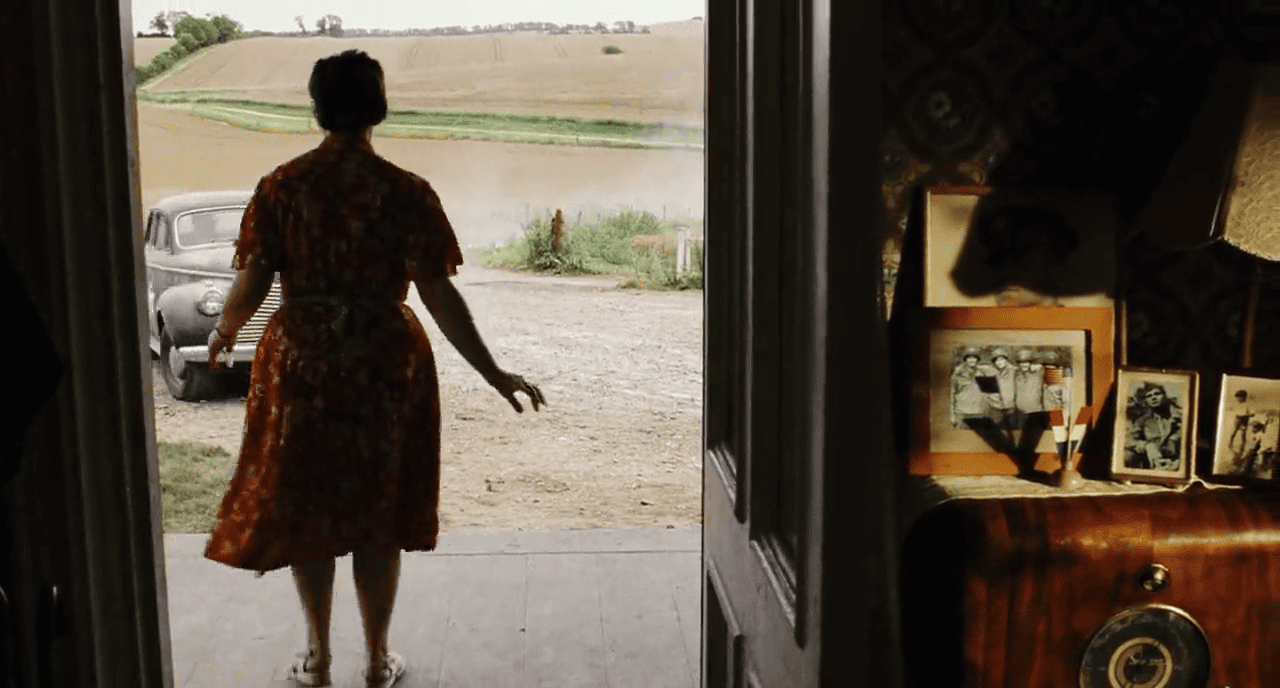}}{\fbox{\parbox[c][2.1cm][c]{.245\linewidth}{\centering Frame 3}}}%
        \hspace{.0065\linewidth}%
        \IfFileExists{figures/Example-4.png}{\includegraphics[width=.245\linewidth]{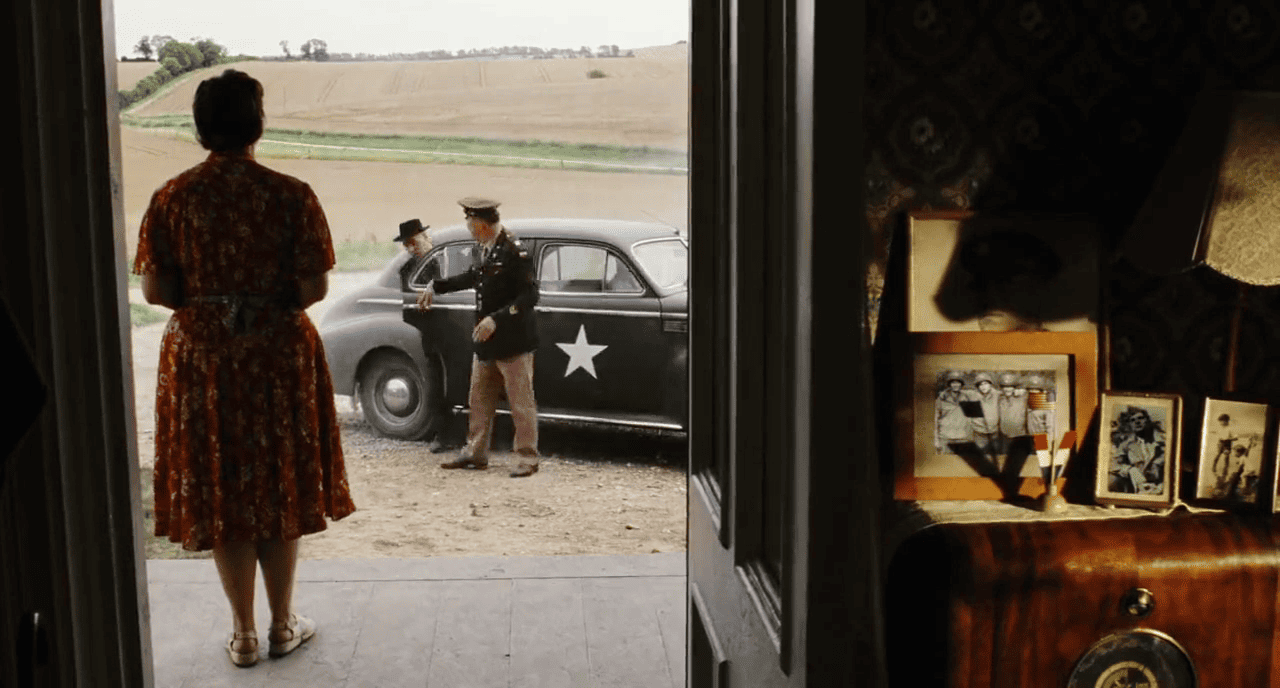}}{\fbox{\parbox[c][2.1cm][c]{.245\linewidth}{\centering Frame 4}}}\\[0.15em]
        {\textbf{Audio:} Composed of cello and string, the overall tempo is relatively slow and the tone is low and deep.}\par
    }
    \vspace{.5em}

    \setlength{\tabcolsep}{2pt}
    \renewcommand{\arraystretch}{1.05}
    \scriptsize
    \begin{tabularx}{\linewidth}{@{}p{2.7cm}|Y|Y|Y@{}}
        \toprule
        \textbf{Method} & \textbf{Audio-Only} & \textbf{Visual-Only} & \textbf{Audio-Visual} \\
        \midrule
        Inputs & Audio \textcolor{ForestGreen}{$\checkmark$}\,\, Video \textcolor{BrickRed}{$\times$} & Audio \textcolor{BrickRed}{$\times$}\,\, Video \textcolor{ForestGreen}{$\checkmark$} & Audio \textcolor{ForestGreen}{$\checkmark$}\,\, Video \textcolor{ForestGreen}{$\checkmark$} \\
        \midrule
        Qwen2.5-Omni (7B) & The audio track seems to be a sad and melancholic piece of music, likely with slow tempo and possibly featuring instruments like strings or \red{piano}. & The video shows a woman in a red dress looking out at a car and a man in a uniform, with a somber and tense emotional tone. & The video shows a woman in a red dress looking out a \red{window} at a man in a black car, with a somber and melancholic musical piece, creating a sense of sadness and longing. \\
        \midrule
        InteractiveOmni (8B) & The song has a melancholic and introspective mood. & The video portrays a woman's journey of acceptance and resilience as she faces the return of \red{her husband from war}, symbolizing the enduring power of \red{love and hope}. & \red{The song} has a melancholic and introspective mood, with a slow tempo and minor key. \\
        \midrule
        VideoLLaMA2 (7B) & \red{The audio track is of poor quality and the voice is inaudible.} & The video shows a woman standing in front of a \red{window}, and a man standing outside a car parked in front of her home, suggesting a somber or serious mood. & The video shows a woman standing in front of a \red{window} looking out, while the audio features \red{a man talking to her}. The overall emotion of the scene is sadness and longing. \\
        \midrule
        Emotion-LLaMA & In the audio, \red{a woman speaks in a normal tone}, expressing a sense of sadness and disappointment. & The woman's \red{facial expression}, which is a mix of sadness and \red{fear}, suggests that she is experiencing emotional distress. & The woman in the image has a sad expression, with \red{her eyes cast downward and a furrowed brow}, indicating a sense of \red{ disappointment}. \\
        \midrule
        AffectGPT & Based on the audio track's description of sound characteristics, it can be inferred that the individual is experiencing sadness, contemplation, and nostalgia. & The \red{man's serious facial expression} and body language, imply that he might be experiencing deep thought, contemplation, or sorrowful emotions. & The \red{man's facial expressions hints contemplation}, along with the melancholic melody of the \red{violin}, the man might be experiencing a sad emotion. \\
        \midrule
        \rowcolor{oursbg}Ours & The audio clip captures a \textcolor{blue}{somber and reflective} mood, with \textcolor{blue}{soothing and low strings and cello} together creating a \textcolor{blue}{depressing and subdued} atmosphere. & A woman stands in a doorway as a black car pulls up and two uniformed men step out. The darkness of the surroundings feels \textcolor{blue}{oppressive}, as if something significant is about to happen. & A woman stands in the doorway to greet the men in the car, seen from behind to heighten the \textcolor{blue}{mystery}. \textcolor{blue}{Low cello and string} add an \textcolor{blue}{sadness}, giving the scene an overall \textcolor{blue}{sorrowful} mood. \\
        \bottomrule
    \end{tabularx}
    \endgroup
\end{table*}

\section{Experiments}
\label{sec:experiments}

\subsection{Benchmark and Baselines}

We organize experiments around two questions that mirror our two-stage design: (1) does Stage 1 with VG-Align enable a strong VLM to hear, and (2) does Stage 2 with EmoAdapter strengthen cross-modal emotion understanding in artworks? For hearing ability, we adopt the Audio Instruction Benchmark (AIR-Bench) \cite{yang2024air}, which spans speech, acoustic, and music understanding. For artistic emotion, we propose the Art-centric Emotion Benchmark (ArtEmoBenchmark), built from movie clips with background music and designed to evaluate content and emotion understanding under three input modalities (audio-only, visual-only, and audio-visual joint).

Baselines cover ALMs, VLMs, and AVLMs. ALMs include Qwen-Audio \cite{chu2023qwen} and Qwen2-Audio \cite{chu2024qwen2}, both trained on more than 100k hours of audio data. VLMs include Qwen2.5-VL~\cite{bai2025qwen2_5vl} and InternVL2.5~\cite{chen2024expanding}, which are strong on visual understanding and language generation. AVLMs include Qwen2.5-Omni~\cite{xu2025qwen2}, InteractiveOmni~\cite{tong2025interactiveomni}, and VideoLLaMA2~\cite{cheng2024videollama} trained on large-scale audio-visual data.
Furthermore, we include two emotion-focused AVLMs, Emotion-LLaMA~\cite{cheng2024emotion} and AffectGPT~\cite{lian2025affectgpt}, which are trained on emotion understanding datasets and specialize in emotion recognition and understanding.
Notably, on AIR-Bench we group the baselines by audio pretraining dataset size, split into large-scale with more than 100k hours and small-scale with fewer than 100k hours.
To match our training size of approximately 70.5 hours, we also include small-scale ALMs such as Next-GPT~\cite{wu2024next}, BLSP~\cite{wang2023blsp}, and SpeechGPT~\cite{zhang2023speechgpt}, making the comparison fairer.
We standardize evaluation across models for fairness: unless a model requires otherwise, we use the same prompts, greedy decoding, and identical context formatting.

\subsection{Dataset and Art-centric Emotion Benchmark}

\textbf{Datasets.} We use different datasets for the two stages to match their different objectives. For Stage 1, which teaches the VLM to hear by seeing, we subsample 70.5 hours samples from the VGGSound dataset~\cite{chen2020vggsound}, which contains audio-visual correspondent videos. The subset covers diverse categories including environmental sounds, instruments, and human voices. By training with these synchronized clips, our VG-Align transfers perception from the visual pathway to an audio pathway with minimal audio hours.
For Stage 2, which strengthens cross-modal emotion understanding, we produce an art-centric dataset composed of 30-second movie clips whose background music is carefully curated to match the visual content and amplify the emotional atmosphere.
From these clips we construct 16720 instruction-following question-answer pairs that include both content understanding and emotion understanding, enabling the model to connect narrative elements with emotional intent. Because our goal centers on emotion, we annotate emotion labels (valence and arousal) for each clip under three input modalities (audio-only, visual-only, and audio-visual joint), so the model can learn how emotion changes with modality. To mitigate catastrophic forgetting of general audio-visual understanding, we further add 8325 samples filtered from VideoInstruct100K with durations between 10 and 30 seconds.

\textbf{Dataset annotation.} Annotation follows a consistent protocol. We collect 1432 movie clips from web sources, each paired with background music selected by the original creators to enhance emotional impact. Then we invite five annotators to annotate emotion labels under three modalities (audio-only, visual-only, and audio-visual joint) using the classic Valence-Arousal model for each clip. Based on these clips and emotion labels, we use the \emph{OpenAI o3-mini} model to generate instruction-following question-answer pairs that cover content and emotion questions, yielding 16720 training samples for Stage 2.

\textbf{Art-centric Emotion Benchmark.} To evaluate artistic emotion understanding, we build the first art-centric emotion benchmark (ArtEmoBenchmark) with 1200 multiple-choice questions across audio-only, visual-only, and audio-visual joint inputs. We first prompt Qwen2.5-Omni to produce textual descriptions of music and visual content and their emotions, then prompt GPT-4o to synthesize multiple-choice questions from these descriptions. Each question has four options with one correct answer.
In the single-modality setting we include four tasks: overall content understanding, overall emotion understanding, specific content questions, and specific emotion questions. Under audio-visual joint input we include overall emotion understanding and specific questions partitioned by audio-centric, video-centric, and cross-modal focus. Overall refers to a direct query about the global content or emotion of the clip, while specific targets concrete details (e.g., a specific instrument's emotion expression).
Furthermore, after automated generation, all questions are manually verified to ensure quality: we invite five annotators to review the questions and options for accuracy and reasonableness, correcting any issues to guarantee the benchmark's integrity.

\subsection{Have VLMs Learned to Hear?}
In Stage 1, our goal is to teach the VLM to hear; therefore we need to evaluate the model's audio understanding ability. As shown in Tab.~\ref{tab:air_benchmark}, we compare our model with various baselines on AIR-Bench. As our training data size is approximately 70.5 hours, we mainly compare with small-scale audio LLMs that are trained with fewer than 100k hours of audio data.
Our model averages 35.17\%, outperforming the best prior average 31.50\% by 3.67 points in small-scale audio LLMs. Replacing response-level logit alignment with feature alignment reduces the average to 31.59\%, 3.58 points below our full Stage 1 and only 0.09 points above the strongest prior audio LLMs, which matches our design intuition that aligning next-token distributions transfers reasoning more faithfully than forcing feature coincidence.
The largest improvements emerge on compositional and music understanding where cross-token context is critical, for example Sound QA 49.90\% versus 44.25\% with feature alignment and Music QA 57.62\% versus 50.12\%.
These gains likely arise because such tasks benefit more from vision, which VG-Align leverages to enhance audio perception.
These results verify that VG-Align can teach the VLM to hear while remaining data-efficient.

\subsection{Can LLMs Understand Artistic Emotion?}
In Stage 2, our goal is to enhance cross-modal emotion understanding for artworks. Therefore, we evaluate on ArtEmoBenchmark and compare with audio-only, visual-only, and audio-visual baselines.
As shown in Tab.~\ref{tab:emotion_benchmark}, our model achieves the best performance across all modalities. The overall average reaches 72.8\% versus 63.0\% for the best baseline, a gain of 9.8 points. Single-modality strength remains high, with 77.0\% on audio-only (+8.0 points) and 87.8\% on video-only (+4.8 points). Crucially, joint input rises to 53.5\% (+11.5 points). The largest advantages appear on cross-modal specific questions, while audio-centric and video-centric queries also benefit. The joint improvement surpasses either modality alone, indicating that EmoAdapter learns cross-modal emotion understanding rather than favoring one modality.
Our model also exceeds single-modality specialists in their strongest modalities, outperforming Qwen2-Audio on audio-only and InternVL2.5 on video-only, which shows that Stage 1 provides solid audio perception while Stage 2 converts complementary cues into measurable emotion gains. Overall, these results confirm that the Emotion Enhancer and Emotion Supervisor in our framework improve artistic emotion understanding beyond either modality.

\subsection{What Drives the Gains?}
We have verified that each stage meets its design goal, so we now ablate key components to clarify their contributions.
The ablations in Tab.~\ref{tab:ablation} clarify how each component contributes to artistic emotion understanding.
An Audio Adapter alone lets the VLM hear, but offers limited audio ability at 29.5\% while keeping video-only near 72.3\% as the Audio Adapter does not affect the visual pathway.
Adding the LoRA raises audio-only to 45.8\% and enables joint input at 37.5\%, yielding an overall 51.8\%. Introducing the Emotion Enhancer lifts all three modalities to 52.3\% (audio), 76.8\% (video), and 43.8\% (joint), improving the overall average by 5.8 points over LoRA alone. Replacing the enhancer with the Emotion Supervisor yields an overall 52.7\%, a smaller gain.
Combining enhancer and supervisor with LoRA and the Audio Adapter delivers the full model at 77.0\% (audio), 87.8\% (video), and 53.5\% (joint), for an overall 72.8\%. This final jump amounts to +15.2 points over enhancer-only and +20.1 points over supervisor-only, showing complementarity and aligning with our design: Stage 1 imparts audio perception efficiently, and Stage 2 adds emotion-focused residuals and emotion supervision that unlock cross-modal and further single-modality gains.

\subsection{Qualitative Analysis}
As shown in Tab.~\ref{tab:qualitative_av_responses}, AVLM baselines are evaluated under audio-only, visual-only, and audio-visual inputs on a Saving Private Ryan scene where officers deliver a death notice.
Despite accepting both modalities, the baselines misread modality-specific cues under restricted inputs. In audio-only, they hallucinate instruments or speech: Qwen2.5-Omni adds piano to low strings and VideoLLaMA2 claims a voice. In visual-only, they infer storylines not supported by frames, such as InteractiveOmni narrating the return of a husband from war or anchoring on a window. Under joint input, inconsistencies remain, with VideoLLaMA2 stating that a man is talking and InteractiveOmni describing the song.
Emotion-focused AVLMs do not fix this trend. Emotion-LLaMA asserts that a woman speaks in the audio and over-attends to local facial expressions, and AffectGPT attributes emotion to a man's expressions and a violin, both underweighting overall scene composition, pacing, and soundtrack, which limits holistic affect perception.
Our model instead aligns the sustained low strings and cello with the somber staging and uniforms to yield a coherent scene-level sadness under all inputs. This advantage reflects our two-stage design that guides fusion toward global affective cues rather than incidental details.

\section{Conclusion}
We present VAEmotionLLM, a vision-anchored audio-visual LLM that learns to hear by seeing and to understand artistic emotion. Our two-stage design couples Vision-Guided Audio Alignment (VG-Align), which transfers perception to an audio pathway by matching response distributions on synchronized clips under shared positional encodings, with a lightweight Cross-Modal Emotion Adapter (EmoAdapter) that injects emotion-sensitive residuals and applies explicit emotion supervision over audio, visual, and audio-visual joint inputs while keeping the backbone largely frozen through parameter-efficient tuning. On the Art-centric Emotion Benchmark (ArtEmoBenchmark) built from movies with background music, the model consistently outperforms audio-only, visual-only, and audio-visual baselines. In the future, we will explore longer temporal reasoning and richer, broader artistic genres.

{
    \small
    \bibliographystyle{ieeenat_fullname}
    \bibliography{main}
}

\end{document}